\documentclass[journal]{IEEEtran}

\ifCLASSINFOpdf
\else
\fi

\usepackage{wrapfig,lipsum,booktabs}
\usepackage{float}
\usepackage{graphicx}
\usepackage{caption}
\usepackage{subcaption}
\usepackage{pifont}
\usepackage{amsmath,amssymb}
\usepackage{hyperref}
\def\l{\lambda}
\def\h{{\mathbf h}}
\def\x{{\mathbf x}}
\def\y{{\mathbf y}}

\def\z{{\mathbf z}}
\def\m{{\mathbf m}}

\def\D{{\mathcal D}}
\def\L{{\mathcal L}}

\def\mix{\text{mix}}

\def\u{{\mathbf u}}
\def\v{{\mathbf v}}

\newcommand{\E}{\mathbb{E}}
\def\vg{{\mathbf{g}}}
\def\gB{{\mathcal{B}}}
\newcommand{\Var}{\mathrm{Var}}

\usepackage{amsmath}
\usepackage{amssymb}
\usepackage{mathtools}
\usepackage{amsthm}

\usepackage{booktabs}
\usepackage{multirow}

\usepackage{mathrsfs}
\usepackage{makecell}

\theoremstyle{plain}
\newtheorem{proposition}{Proposition}[section]

\theoremstyle{definition}
\theoremstyle{remark}
\usepackage{pifont}
\usepackage{amsmath,amssymb}

\usepackage{xcolor}

\usepackage{multicol}

\begin{document}
\title{{Mixup Augmentation with Multiple Interpolations}}

\author{Lifeng~Shen,
        Jincheng~Yu,
        Hansi~Yang,
        and~James~T.~Kwok,~\IEEEmembership{Fellow,~IEEE}
\thanks{Lifeng Shen is with the Division of Emerging Interdisciplinary Areas (Artificial Intelligence), The Hong Kong University of Science and Technology,
Hong Kong, China. (e-mail: lshenae@connect.ust.hk).}
\thanks{Jincheng~Yu, Hansi~Yang, James T. Kwok are with the Department of Computer Science and Engineering, The Hong Kong University of Science and Technology,
Hong Kong, China. (e-mail: jincheng.yu@connect.ust.hk, {hyangbw,jamesk}@cse.ust.hk)}
}

\maketitle

\begin{abstract}
Mixup and its variants form a popular class of data augmentation techniques.
Using a random sample pair,
it generates a new sample
by linear interpolation of the inputs and labels.
However, generating only one single interpolation
may limit its augmentation
ability.
In this paper, we propose a simple yet effective extension called multi-mix, which
generates multiple interpolations
from a sample pair.
With an ordered
sequence of generated samples, multi-mix can better guide the
training process than standard mixup. Moreover,
theoretically, this can also reduce the
stochastic gradient
variance.
Extensive experiments
on a number of synthetic and large-scale data sets
demonstrate
that multi-mix outperforms
various mixup variants and non-mixup-based baselines
in terms of
generalization, robustness, and calibration.
\end{abstract}

\begin{IEEEkeywords}
mixup, data augmentation, deep learning.
\end{IEEEkeywords}

\IEEEpeerreviewmaketitle

\section{Introduction}

In recent years,
deep networks have made significant breakthroughs in fields
such as computer vision
and natural
language processing.
However, deep networks are often large, data-hungry and can easily overfit,
especially in the presence of adversarial examples \cite{shwartz2017opening} or
distributional difference between the training and test data sets \cite{ben2010theory,pmlr-v37-long15}.
Hence, an important issue is how
to improve the generalization of deep networks.

\begin{figure}[t!]
\centering
\includegraphics[width=0.33\textwidth]{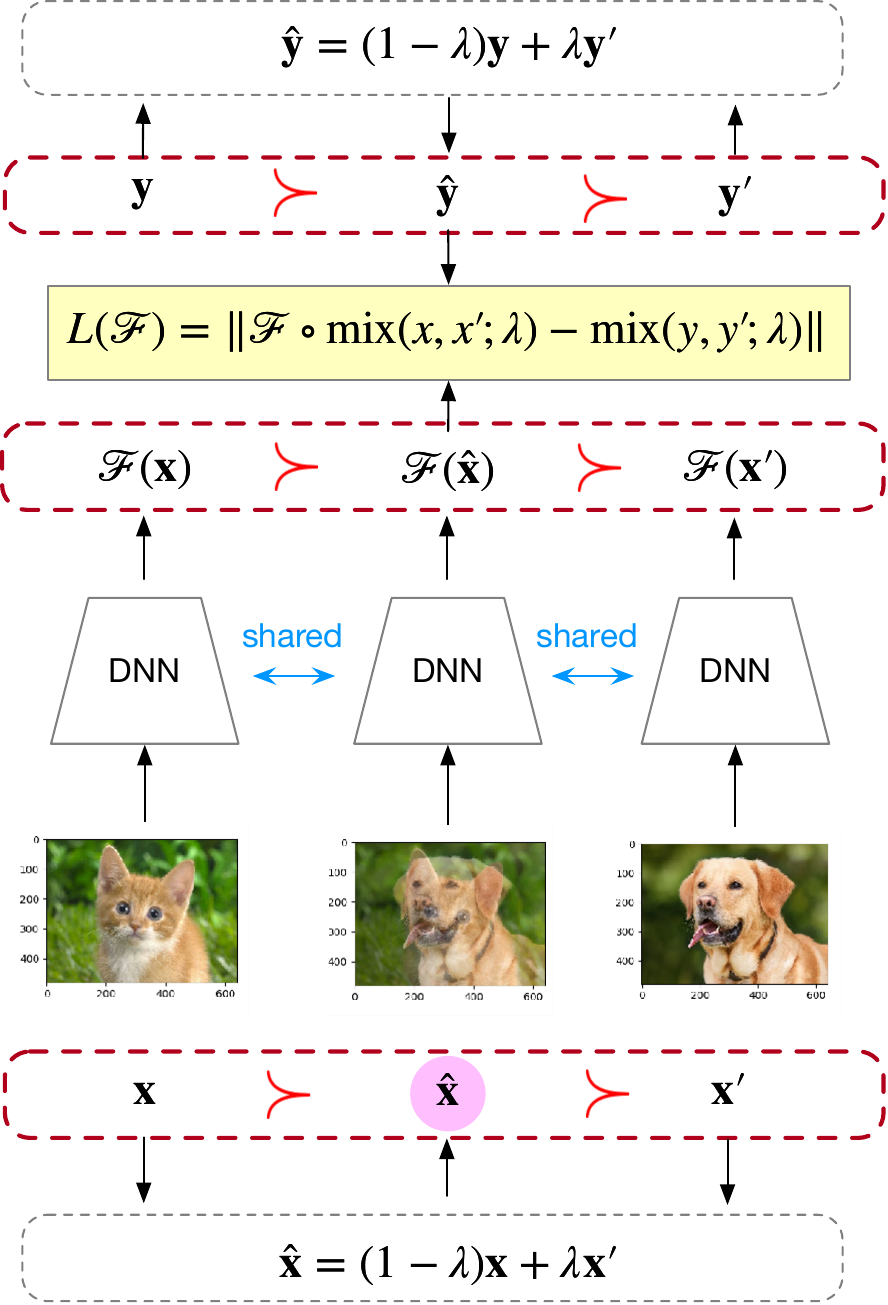}
\caption{An example of mixup training. Here, a cat and a dog are mixed with a
mixing coefficient $\lambda=0.5$. The symbol $\succ$ describes relative
ordering along the mixup transformation paths (cat $\succ$ (half-cat+half-dog) $\succ$ dog) in both input and output spaces. Mixup training improves performance by leveraging order relationship
along with mixup transformation paths in the hidden space of $\mathcal{F}$.}
\label{fig:concept_mixup}
\end{figure}

Data augmentation \cite{shorten2019survey}
has been widely used to enhance the performance and generalization capabilities of deep learning models. It enlarges the given training set by applying a variety of transformations to the provided training samples while preserving the original
data semantics.
Common transformations for image data include random rotation, translation, scaling, and flipping.
By introducing these variations, data augmentation helps to mitigate overfitting by providing the model with a more diverse and representative set of examples.

In recent years, mixup \cite{zhang2018mixup} has emerged as a powerful data augmentation technique in deep learning. It generates augmented training samples by linearly interpolating pairs of input examples and their corresponding labels. 
Specifically,
given an input-label
sample pair $\{(\x, \y), (\x',\y')\}$ from the
training set $\D$,
mixup
creates a new sample
$(\hat{\x},\hat{\y})$
via interpolation
(Figure \ref{fig:concept_mixup}), where: 
\begin{equation} \label{eq:mixup}
\hat{\x}=\mix(\x, \x'; \lambda), \;\; \hat{\y}=\mix(\y, \y'; \lambda),
\end{equation}
$\lambda$ is a
mixing coefficient
drawn from the Beta distribution
$Beta(\alpha, \alpha)$ for some $\alpha\in(0, \infty)$, and
$\mix(\cdot;\cdot;\lambda)$ is a linear interpolation that mixes the
two input arguments based on
$\lambda$:
\begin{align}
\mix(\z,\z';\lambda)=(1-\lambda)\z+\lambda\z'.
\label{eq:interpolation}
\end{align}
Using the implicit order in both the input (denoted $\x\succ\hat{\x}\succ\x'$) and output ($\y\succ\hat{\y}\succ\y'$),
the deep network
$\mathcal{F}$
is encouraged
to behave linearly in-between the training examples
$\{(\x, \y), (\x',\y')\}$.
In other words,
we expect a similar ordering on
the network outputs:
$\mathcal{F}(\x)\succ\mathcal{F}(\hat{\x})\succ\mathcal{F}(\x')$.

Mixup has shown promising results in improving model performance, enhancing generalization, and reducing overfitting
\cite{zhang2018mixup}.
Following its success, a number of variants have been proposed.
For instance, manifold mixup \cite{ManifoldMIXUP} explores the application of
mixup in the hidden space. FMix \cite{harris2020fmix} incorporates mixing masks of arbitrary shapes, enabling more flexible and diverse sample mixing. Puzzle-mix \cite{kim2020puzzle} and co-mixup \cite{kim2021comixup} leverage saliency information to identify informative mixup policies, further enhancing the quality and informativeness of the mixup process.
Although
mixup and its variants show
improved
performance, 
their use of only one single interpolation from each sample pair may
limit its augmentation ability.

In this paper, we propose  a simple mixup extension called
{multi-mix},
which generates multiple interpolation samples for each sample (and label) pair.
For example,
from two source samples $\x$ and $\x'$,
multi-mix
can generate three interpolations
$\hat{\x}_1,\hat{\x}_2,\hat{\x}_3$ with
$\x\succ\hat{\x}_1\succ\hat{\x}_2\succ\hat{\x}_3\succ\x'$.
By using an ordered sequence of mixup samples,
{multi-mix} can better
guide the training process than standard mixup.
Moreover,
it can be shown theoretically that
it also reduces
the
variance
of the stochastic gradient.
Extensive experiments are performed to compare the proposed multi-mix with a
number of baselines
on both
synthetic and large-scale data sets. Results demonstrate that multi-mix
has superior performance in various aspects including generalization, robustness, and calibration.

The rest of this paper is organized as follows.
Section \ref{sec:relatedwork}
reviews the related work on the standard mixup and its popular variants.
The proposed multi-mix algorithm and corresponding theoretical analysis are presented in Section \ref{sec:method}.
Experimental
results are reported in Section \ref{sec:experiment}, and the last section gives
some concluding remarks.

\begin{figure*}[t!]
\centering
\includegraphics[width=0.9\textwidth]{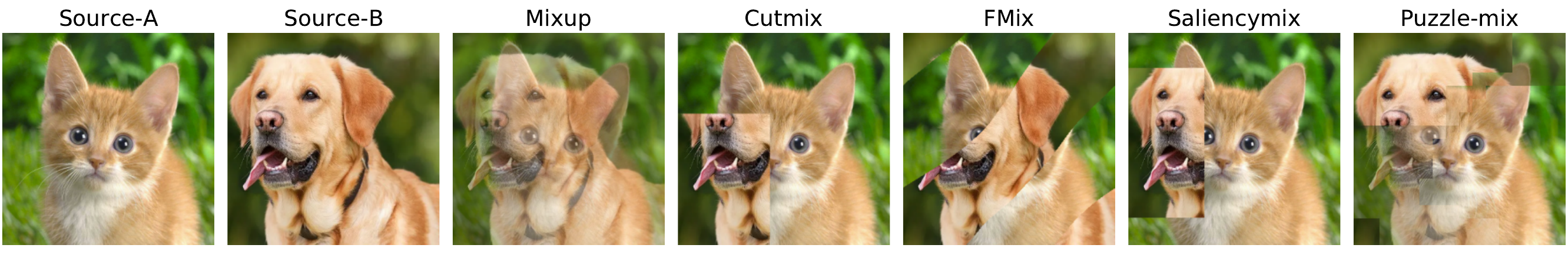}
\caption{Examples of input mixup and saliency-based mixup techniques.}
\label{fig:mixup_example}
\end{figure*}

\section{Related Work:
Mixup and its Variants}\label{sec:relatedwork}

Mixup \cite{zhang2018mixup} is a recent augmentation approach based on linear interpolation.
Due to its simplicity, mixup has attracted much interest in recent
years
\cite{yun2019cutmix,harris2020fmix,ManifoldMIXUP,uddin2020saliencymix,kim2020puzzle,kim2021comixup,venkataramanan2022alignmixup}.
The various mixup-based methods can be roughly divided into three categories: (i) input mixup,
(ii) manifold mixup, and (iii) saliency-based mixup.

Input mixup \cite{zhang2018mixup} and its variants
\cite{yun2019cutmix,harris2020fmix,hong2021stylemix,yang2022recursivemix}
augment the data by directly manipulating the input.
Besides the standard input mixup, a well-known variant is cutmix
\cite{yun2019cutmix}.
Given two training images $\x, \x'$, and the corresponding labels $\y, \y'$,
cutmix
obtains an augmented sample
$(\hat{\x},\hat{\y})$ as:
\begin{align}
\hat{\x}&=(\mathbf{1} -\mathbb{I}_{B})\odot\x+\mathbb{I}_{B}\odot\x', \;\; \hat{\y}=\mix(\y, \y'; \lambda),
\label{eq:cutmix}
\end{align}
where
$\mathbb{I}_{B}$ is a randomly-sampled binary rectangular mask, and
$\odot$ is the element-wise product. The mixing ratio $\lambda$
is
calculated as the ratio of the area of the rectangular mask
and the size of the whole mask $\mathbb{I}_{B}$.
FMix \cite{harris2020fmix} further
extends cutmix  by using
binary masks of
arbitrary shapes. In Stylemix \cite{hong2021stylemix}, the content and style
information of the input image pairs are leveraged for mixing.
Recursivemix \cite{yang2022recursivemix} explores a recursive mixed-sample learning
paradigm in the input space to improve mixup training.

Instead of performing interpolation only at the input, manifold mixup
\cite{ManifoldMIXUP} allows interpolation at some
hidden
layer
in a deep network.
For each data batch,
a layer $s$ (with
features $h^s(\cdot)$)
is first randomly sampled.
As in (\ref{eq:mixup}), an augmented sample
$(\hat{\h}^s,\hat{\y})$
is generated by interpolating
$(h^s(\x), \y)$
and $(h^s(\x'), \y')$:
\begin{align}
\hat{\h}^s&=\mix(h^s(\x), h^s(\x'); \lambda), \;\; \hat{\y}=\mix(\y, \y'; \lambda).
\label{eq:manifold_mixup}
\end{align}
Empirically, this induces smoother boundaries and better generalization.
When $s=0$ (the input layer),
manifold mixup reduces to standard input mixup.
In AlignMixup \cite{venkataramanan2022alignmixup}, a matching objective in the hidden space is developed for feature alignment.
However, AlignMixup is computationally more expensive as it introduces an additional autoencoder during mixup training.

Saliency-based mixup methods \cite{uddin2020saliencymix,kim2020puzzle,kim2021comixup,SageMix,BoostingMix,SAGE}
leverage the data's saliency information to find an informative mixup policy, and
have
shown
performance
superior
to input and manifold mixup.
Instead of using a random selection strategy for mixing,
saliencymix \cite{uddin2020saliencymix} 
selects an informative image
patch with a saliency map.
Puzzle-mix \cite{kim2020puzzle} 
exploits saliency and local statistics to learn an 
interpolation
strategy for mixing.
The augmented sample
$(\hat{\x}, \hat{\y})$
is obtained
as:
\begin{eqnarray}
\hat{\x}&= &(1-\m)\odot\Pi_0^\top s(\x)+\m\odot\Pi_1^\top s(\x'),
\label{eq:puzzlemix}\\
\hat{\y} & = & \mix(\y, \y'; \hat{\l}),
\nonumber
\end{eqnarray}
where
$s(\cdot)$
measures the saliency,
and $\Pi_0$ and $\Pi_1$ are
learnable transport plans for the
inputs.
The mask $\m$,
with elements $m_i$'s in $[0,1]$,
is represented as a
vector.
Each $m_i$ is
discretized to
$d+1$ values:
$\{\frac{t}{d} \;|\; t=0,1,\dots,d\}$.
Given a $\l$ that is randomly sampled from a Beta distribution, $m_i$ is
generated from the mixing prior
$p\left(m_i=\frac{t}{d} \right)=\binom{d}{t}\l^t(1-\l)^{d-t}$.
Consequently, the mixing ratio $\hat{\l}$ for label mixing
is set as $\frac{1}{n}\sum^n_{i=1} m_i$.
Intuitively, the mask $\m$ and transport plans try to maximize the saliency of the
revealed portion of the image.
Co-mixup \cite{kim2021comixup} is
an extension of puzzle-mix. 
Instead of mixing only a random pair of input samples,
co-mixup
generates a batch of mixup samples
by accumulating many salient regions from multiple input samples.

Figure \ref{fig:mixup_example}
shows examples generated by
input mixup and some of its 
variants.
As can be seen, input mixup simply overlaps the two source images, which weakens the original information of the object.
Cutmix and FMix can
well preserve the clarity of the original object. However,
the most informative parts may not be included in the mixing
result.
Saliencymix
well covers the important area, but may have sharp borders in the resulting image.
The image by puzzle-mix covers the most informative parts and
also has smoother borders.
This suggests that choosing puzzle-mix as the basic mixup operation in multi-mix is more effective in introducing additional information about the mixup transformation path, thereby aiding in the regularization of network training.

\begin{figure}[t!]
  \begin{center}
  \includegraphics[width=0.9\linewidth]{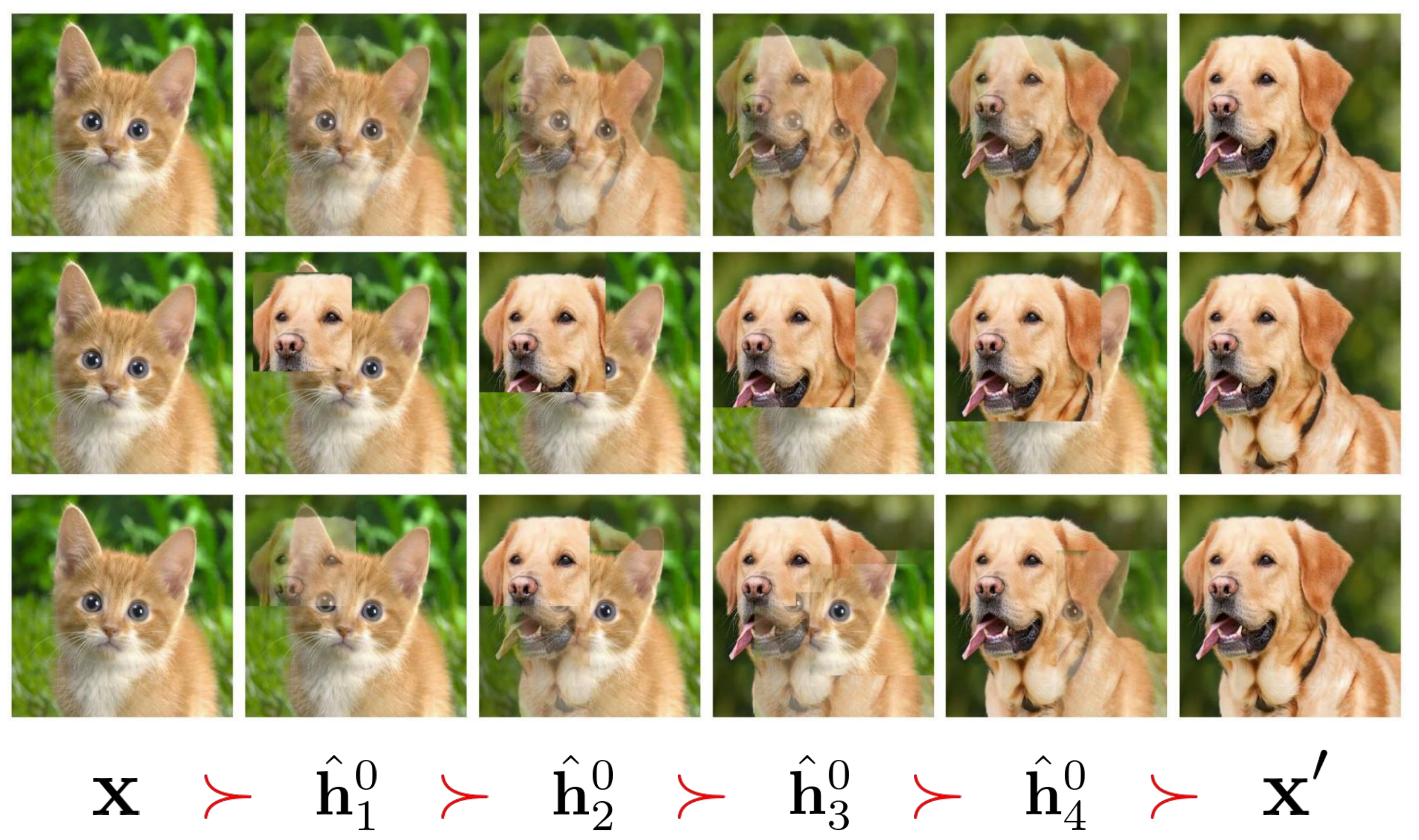}
  \end{center}
  \caption{Examples interpolations ($\hat{h}^0_1,
\hat{h}^0_2,
\hat{h}^0_3,
\hat{h}^0_4$) generated by
input mixup (top), cutmix
(middle), and puzzle-mix (bottom).}
  \label{fig:multimix}
\end{figure}

\section{Multi-Mix: Mixup with Multiple Interpolations}
\label{sec:method}

Existing mixup methods generate only one single interpolation for each sample
pair.
To introduce more mixup information
on the mixup transformation path  for regularizing the network,
we propose {multi-mix}, which
generates mixup sample pairs
$\{(\hat{\h}_k^s, \hat{\h}_k^s)\}_{k=1}^K$ with
$K>1$
interpolations. 

Section~\ref{sec:model}
describes how this can be extended to multiple interpolations.
Section~\ref{sec:theory} provides
theoretical analysis on the effect of using multiple interpolations on gradient
variance.
Section~\ref{sec:discuss}
discusses related methods that also
increase the number of mixup samples in each batch.

\subsection{Generating Multiple Interpolations}
\label{sec:model}

In the following,
we focus on
multi-mix
extensions of input mixup, manifold mixup, cutmix and
puzzle-mix.
Extensions for other mixup variants can be developed analogously.

\subsubsection{Extending Input Mixup and Manifold Mixup}

The multi-mix
extensions for input mixup and manifold mixup
are straightforward.
Given a deep network and a pair of random samples $\x$ and $\x'$,
the $K$ interpolations
$\{(\hat{\h}_k^s,\hat{\y}_k)\}_{k=1}^K$
can be generated
from input mixup or manifold mixup
as:
\begin{equation}
\hat{\h}_k^s=\mix(h^s(\x), h^s(\x'); \lambda_k), \; \hat{\y}_k=\mix(\y, \y'; \lambda_k),
\label{eq:manifold_mixup_K}
\end{equation}
where $s$ is the layer index, and
$h^s(\x), h^s(\x')$ are the
layer-$s$
outputs of $\x, \x'$, respectively.
For input mixup, $s=0$; whereas
for manifold mixup, $s>0$.
The
$\l_i$'s
satisfy
$0<\l_1<\dots<\l_K<1$, and are randomly sampled from the same Beta distribution.

\subsubsection{Extending
Cutmix}

For cutmix with $K$ interpolations,
we need to generate
$K$ boxes
indicating the
$K$ regions of $\x$
that are cropped
and replaced by the corresponding regions
in $\x'$.
We first sample
the center
that is shared by the $K$ boxes.
Let its
coordinates be $(r^x, r^y)$,
where
$r^x\sim \mathrm{uniform}(0, W)$,
$r^y\sim \mathrm{uniform}(0, H)$, and
$W, H$ are the image's width and height, respectively.
The
width and height
of the
$k$th box
are
given by
$r^w_k=W\sqrt{{\l}_k}$ and
$r^h_k=H\sqrt{{\l}_k}$, respectively, where
${\l}_k$ is randomly sampled from a Beta distribution.

\subsubsection{Extending
Puzzle-mix}

For puzzle-mix
with $K$ interpolations,
we generate $K$ masks ($\m_k$'s) 
in (\ref{eq:puzzlemix})
by using $K$ mask-mixing priors $\{p_k(m_i)\}_{k=1,2,\dots,K}$.
Each $p_k(m_i)$ is parameterized by a $\l_k$ that is also randomly sampled from a shared Beta distribution.

Figure \ref{fig:multimix}
shows
examples generated by  the multi-mix extensions of
input mixup,
cutmix and
puzzle-mix on
a pair of example
images.
As can be seen,
the
generated
multiple interpolations show a continuous transformation from one image (cat) to another (dog).
For example,
consider the multi-mix extension of cutmix (second row).
When
the cropping box is small,
the augmented sample is closer to the cat (left).
When
the cropping box becomes larger,
the sample becomes more like the dog (right).

\subsection{Variance Reduction with Multi-Mix}
\label{sec:theory}

Theoretical analysis
shows that
a small
variance
in the stochastic gradients
can
lead to faster convergence~\cite{ghadimi2013sgd}.
In this section, we theoretically show that multi-mix can reduce this
gradient
variance.

Instead of only
minimizing the loss on the original data samples,
mixup and its variants
generate
artificial samples
during training and
can be
regarded as optimizing the following {\em mixup
loss}~\cite{carratino2020mixup,zhang2021how}:
\[ \L_{\texttt{mixup}} =\mathop{\mathbb{E}}_{\lambda \sim Beta(\alpha, \alpha)}
\mathop{\mathbb{E}}_{(\x, \y), (\x', \y') \in {\D}} \mathcal{L}((\x,\y), (\x',
\y'),\lambda), \]
where $\D$ is the training set, and
$\mathcal{L}((\x,\y), (\x', \y'),\lambda)$ is the loss on a mixup sample generated
from the sample pair $\{(\x, \y), (\x', \y')\}$ with mixup weight $\lambda$.
Empirically,
for mixup (or its variants) with single interpolation, 
$\L_{\texttt{mixup}}$ is approximated as
$\frac{1}{|{\D}|} \sum_{(\x, \y), (\x', \y') \in {\D}} \mathcal{L}((\x,\y), (\x',
\y'),\lambda)$.

With the proposed multi-mix,
$\L_{\texttt{mixup}}$ is approximated as
\[ \L_{\texttt{multi-mix}} = \frac{1}{K|{\D}|} \sum_{(\x, \y), (\x', \y') \in
{\D}} \sum_{k=1}^K \mathcal{L}((\x,\y), (\x',
\y'),\lambda_k).\]
In a particular training iteration,
a batch $\gB$ of $B$ samples from
${\D}$
are sampled,
and a total of $KB$ mixup samples are generated.
For multi-mix,
the gradient of the mixup loss on $\gB$
is:
\begin{equation}
\tilde{\vg} = \frac{1}{KB} \sum_{(\x,\y), (\x', \y') \in \gB} \sum_{k=1}^K \vg((\x,\y), (\x',
\y'),\lambda_k),\label{eq:grad_multimix}
\end{equation}
where
$\vg((\x,\y), (\x',
\y'),\lambda_k) = \nabla \mathcal{L}((\x,\y), (\x',
\y'),\lambda_k)$ is the loss gradient on a mixup sample generated from
$\{(\x, \y)$, $(\x', \y')\}$ with
mixup weight
$\lambda_k$.
Obviously, we have $\mathbb{E}[\tilde{\vg}] = \nabla \L_{\texttt{mixup}}$.
Define 
\begin{align}
    \Var[\tilde{\vg}] = \mathbb{E}[\|\tilde{\vg} -
	 \mathbb{E}[\tilde{\vg}]\|^2]=\E[\| \tilde{\vg} \|^2]-(\E[\| \tilde{\vg} \|])^2
    \label{eq:gvar}
\end{align}
as
the variance
of 
$\tilde{\vg}$
w.r.t. the data and 
mixup weights
of $\tilde{\vg}$. 
The following Proposition
shows that multi-mix reduces the
gradient's
variance.

\begin{proposition}
$\Var[\tilde{\vg}]$ decreases with $K$.
\label{prop:var1}
\end{proposition}
Proof is in Appendix~\ref{sec:proof}.
While Proposition \ref{prop:var1} suggests the use of a large $K$,
the number of gradient computations also grows with $K$.
Moreover, as will be seen
in
Section \ref{ablation:multiple},
empirically the performance saturates when $K$ is sufficiently large.

\begin{figure*}[t!]
\centering
\subcaptionbox{\label{toy-a} Noisy \textit{Spiral}.}
{\includegraphics[width=0.23\textwidth]{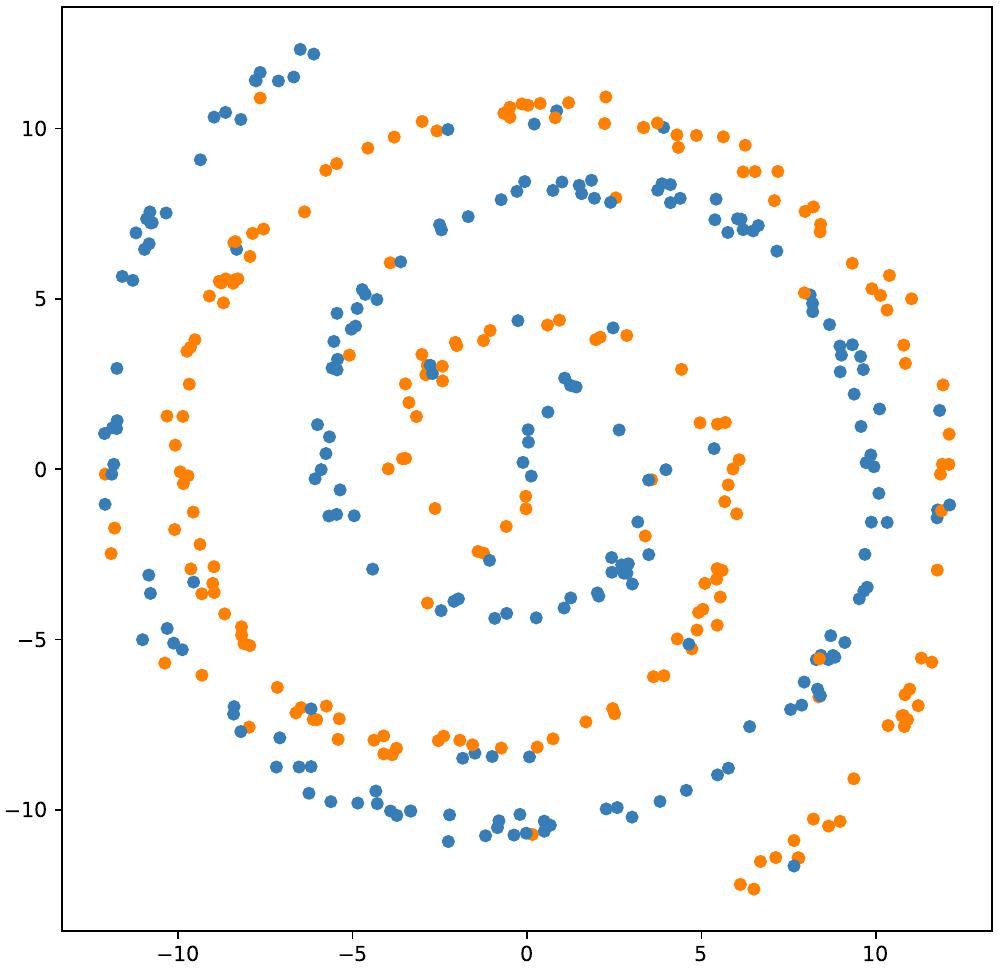}}\hspace{0.2cm}\centering
\subcaptionbox{\label{toy-b}no mixup (90.67\%).}
{\includegraphics[width=0.24\textwidth]{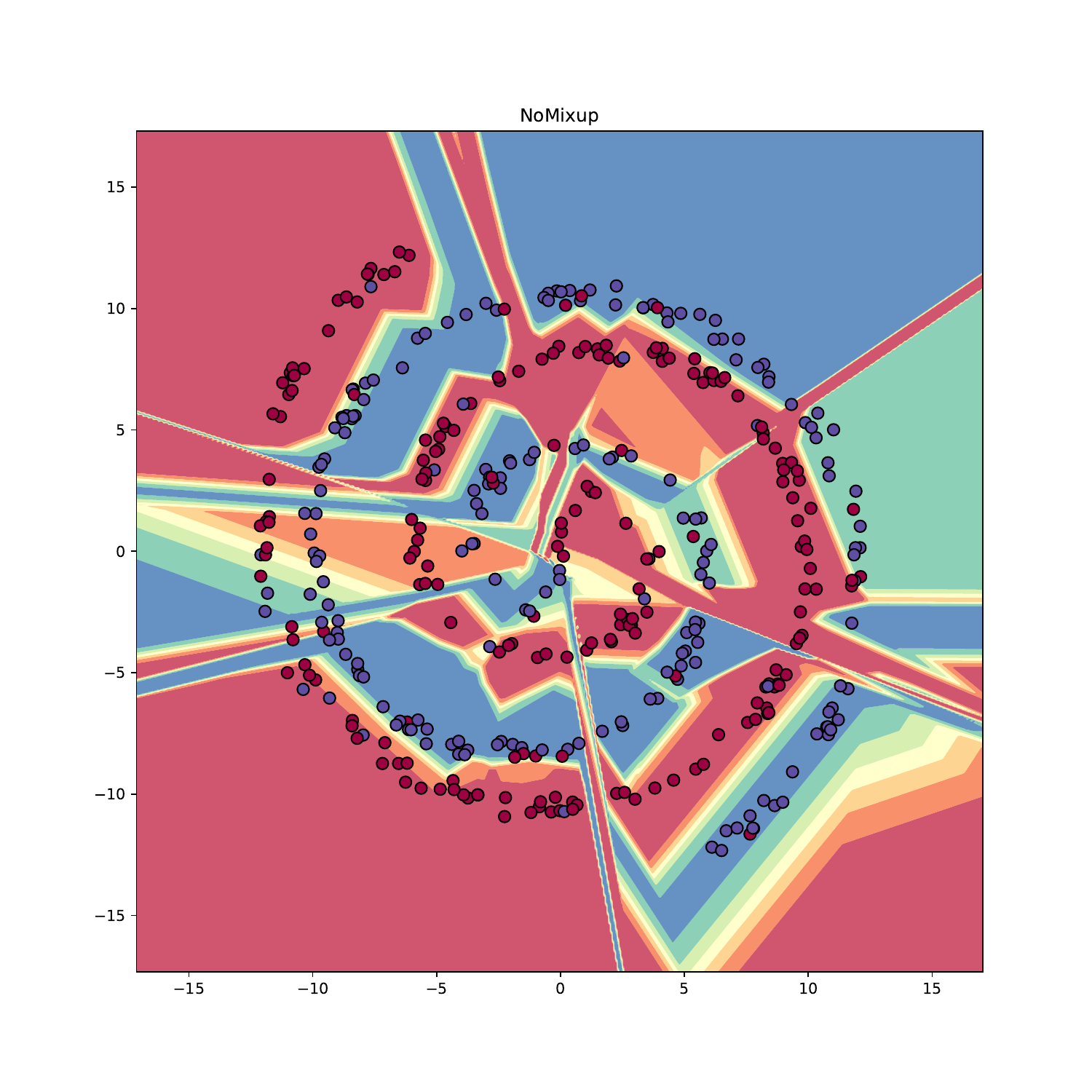}}
\subcaptionbox{\label{toy-c}Manifold-mix (94.17\%).}
{\includegraphics[width=0.24\textwidth]{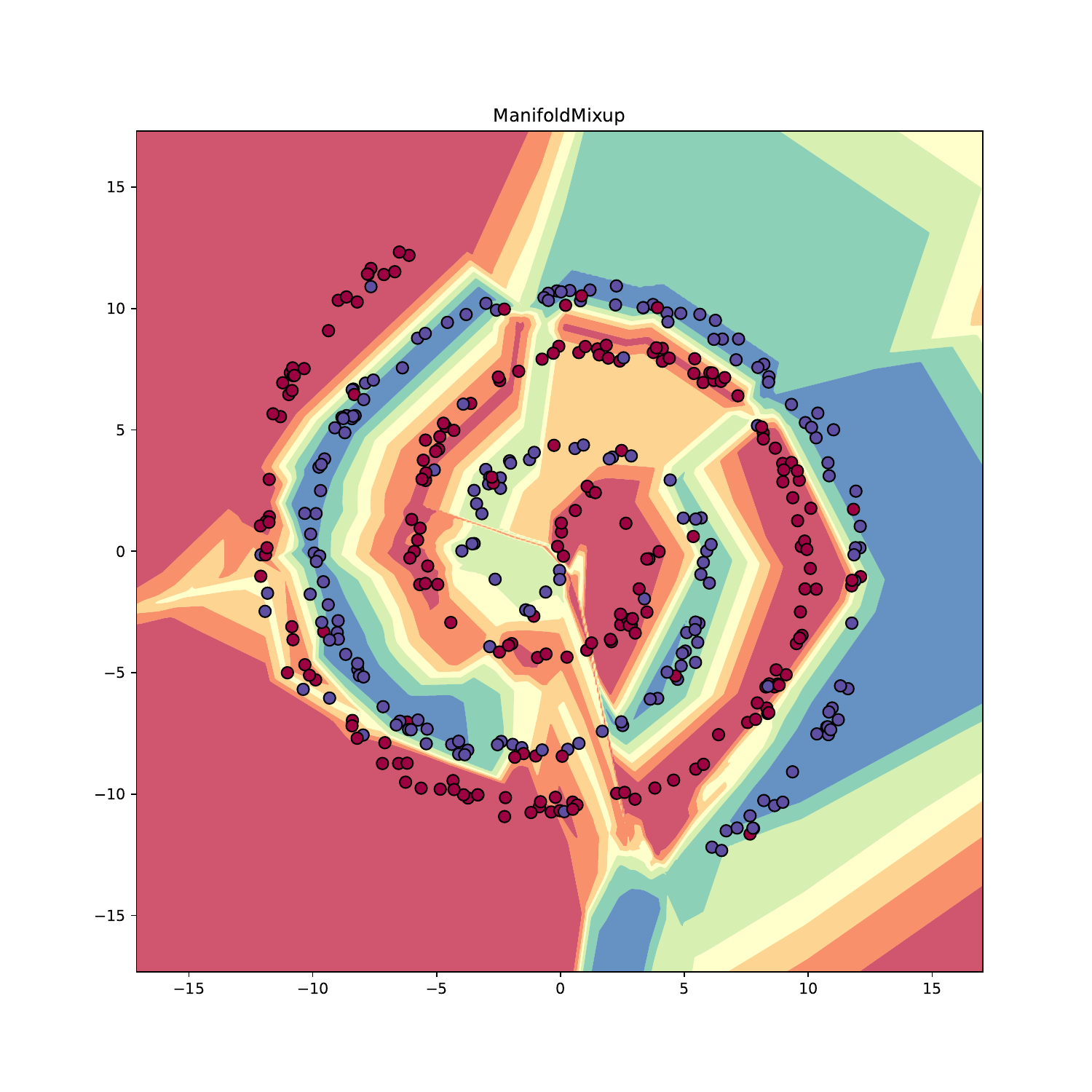}}
\subcaptionbox{\label{toy-d}{Multi-mix (97.50\%).}}
{\includegraphics[width=0.24\textwidth]{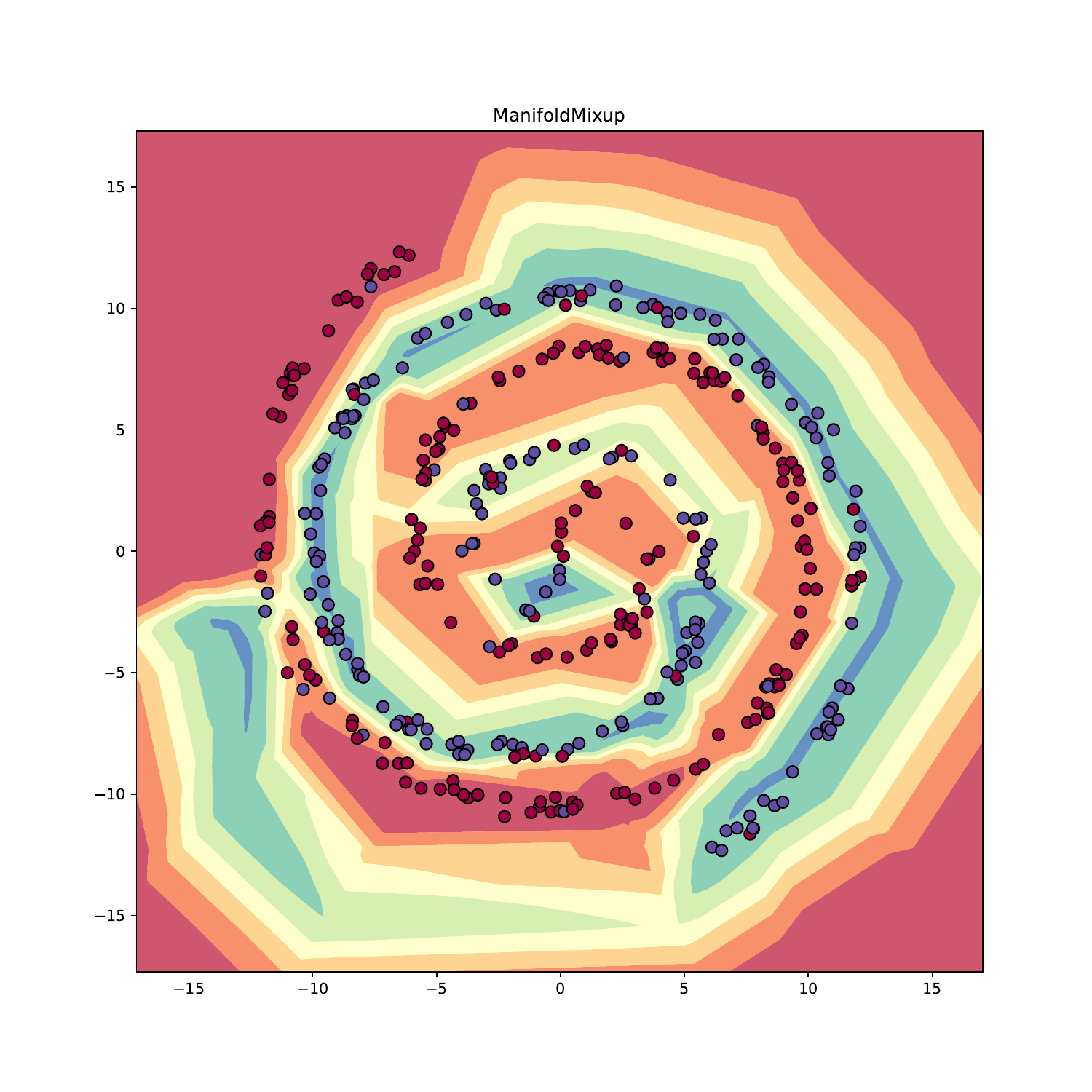}}
\caption{The noisy $\textit{spiral}$ data and decision boundaries learned without
mixup, with manifold mixup, and
manifold mixup
with multiple interpolations.
Values in brackets are the top-1 test accuracies obtained.}
\label{fig:spiral}
\end{figure*}

\subsection{Discussion}
\label{sec:discuss}
The proposed multi-mix introduces more
(i.e., a total of $KB$)
mixup samples
in each batch to help
network training.
In this section, we discuss related methods that also
increase the number of mixup samples in each batch.
Note that they
do not optimize the networks by introducing more mixup transformation path information
as in multi-mix.
As will be empirically demonstrated in Section \ref{sec:experiment},
this mixup
transformation path
is helpful in improving generalization,
robustness against data corruption, transferability, and calibration.

\subsubsection{Large-batch Training}
A simple approach
(often called large-batch training \cite{keskar2017batch})
is to just use a larger batch
size,
without mixup (or its variants).  However, as discussed in \cite{keskar2017batch},
the model's generalization ability
may be degraded
in practice.
This will also be empirically verified in Section~\ref{batch_augmentation}.

\subsubsection{Large-batch Mixup}
Alternatively, one can simply use
standard mixup (or its variants), but with
$KB$ mixup samples. 
To differentiate this from mixup (using only $B$ mixup samples),
we will refer to this as ``large-batch mixup''
in the sequel.
Its loss gradient is
\begin{equation}
\tilde{\vg}' = \frac{1}{KB} \sum_{(\x,\y), (\x', \y') \in \gB} \vg((\x,\y), (\x', \y'), \lambda),\label{eq:grad_unimix}
\end{equation}
where a single $\lambda$ is sampled from $Beta(\alpha, \alpha)$
and used on all sample pairs in the batch.
Note that
both
(\ref{eq:grad_multimix}) and
(\ref{eq:grad_unimix}) involve
$KB$ gradient computations, and thus
have
the same
computational cost in one batch update.

Let $\Var[\tilde{\vg}'] = \mathbb{E}[\|\tilde{\vg}' - \mathbb{E}[\tilde{\vg}']\|]$ be
the variance of the stochastic gradient
$\tilde{\vg}'$
w.r.t.
the data and 
mixup
weights.
The following Proposition
shows that with the same computational cost (i.e., $KB$ gradient computations),
multi-mix has a smaller
variance
than large-batch mixup when $B$ is large enough.
Proof is in Appendix \ref{sec:proof_2}.

\begin{proposition}\label{prop:var}
For any $K > 1$, $\Var[\tilde{\vg}] \le \Var[\tilde{\vg}']$ when
\[ B \ge
\frac{\Var_{(\x,\y), (\x', \y')} \mathbb{E}_{\lambda_k \sim Beta(\alpha, \alpha)}[\vg((\x,\y), (\x', \y'),
\lambda_k)]}{\Var_{\lambda_k\sim Beta(\alpha,\alpha)}\mathbb{E}_{(\x,\y), (\x', \y')}[\vg((\x,\y), (\x', \y'), \lambda_k)]}.
\]
\end{proposition}
With a smaller gradient variance, multi-mix converges faster than
large-batch mixup. 
It also has better generalization performance,
as will be empirically verified in Section~\ref{batch_augmentation}.
Thus, with the same computation cost, using more
interpolations is preferred over the use of more data sample pairs.

\subsubsection{Batch Augmentation}
Multi-mix is also similar to batch augmentation \cite{hoffer2020augment},
which replicates samples in the same batch with different data
augmentations (e.g.,
random cropping and horizontal flipping).
To enhance batch augmentation, mixup can also be used to
further enlarge the dataset.
However, while batch
augmentation focuses on replicating each sample by different transformations,
multi-mix generates multiple mixup samples from each sample pair.

\section{Experiments}\label{sec:experiment}

In this section, we perform extensive evaluation of the proposed multi-mix
 on various
 tasks, including synthetic image classification (Section~\ref{sec:spiral}),
real-world image classification (Sections~\ref{sec:cifar} and \ref{sec:imagenet}),
 weakly-supervised object localization (Section~\ref{sec:wsol}), corruption robustness
 (Section~\ref{sec:corrupt}),
 transfer learning (Section~\ref{sec:xfer}), speech recognition
 (Section~\ref{sec:speech}), and calibration analysis
 (Section~\ref{sec:calib}). 
 Section~\ref{ablation:multiple} 
 presents some
 ablation studies.

\begin{table*}[!ht]
 \centering 
  \renewcommand\arraystretch{1.1}
    \begin{tabular}{c|ccc|c}
    \toprule[1pt]
	 & \multicolumn{3}{c}{{CIFAR-100}} & \multicolumn{1}{|c}{{Tiny-ImageNet}}\\
& \multicolumn{1}{c}{{PreActResNet18}} & \multicolumn{1}{c}{{WRN16-8}} & \multicolumn{1}{c}{{ResNeXt29-4-24}} & \multicolumn{1}{|c}{{PreActResNet18}} \\\toprule
    no mixup & 23.53 $\pm$ 0.39 &  21.93 $\pm$ 0.52 &  21.84 $\pm$ 0.16  & 43.54 $\pm$ 0.13 \\
    Input mixup & 22.49 $\pm$ 0.21 &  20.18 $\pm$ 0.51 &  21.78 $\pm$ 0.09  & 43.35 $\pm$ 0.12  \\
    Manifold mixup & 21.62 $\pm$ 0.29 &  20.49 $\pm$ 0.46 &  21.45 $\pm$ 0.12 & 41.21 $\pm$ 0.10 \\
    Cutout & 21.12 $\pm$ 0.38 & \underline{19.19 $\pm$ 0.39} & 20.86 $\pm$ 0.13 & 40.52 $\pm$ 0.15\\
    Cutmix & 21.29 $\pm$ 0.26 &  20.14 $\pm$ 0.50 &  20.92 $\pm$ 0.10 & 43.11 $\pm$ 0.12 \\
    FMix & 21.52 $\pm$ 0.24 & 19.75 $\pm$ 0.49 & 21.87 $\pm$ 0.10 & 42.38 $\pm$ 0.15 \\
    Co-mixup & \underline{19.89 $\pm$ 0.28} & 19.24 $\pm$ 0.31 & \underline{19.43 $\pm$ 0.08}  & \underline{35.94 $\pm$ 0.11} \\
    Puzzle-mix & 20.61 $\pm$ 0.29 & 19.27 $\pm$ 0.32 & 19.74 $\pm$ 0.09 & 36.79 $\pm$ 0.07 \\
    Label-smoothing & 23.09 $\pm$ 0.34 & 21.45 $\pm$ 0.44 & 20.66 $\pm$ 0.12 & 43.04 $\pm$ 0.12\\
    Multi-mix (with
Puzzle-mix) & \textbf{19.19} $\pm$ \textbf{0.27} & \textbf{18.52} $\pm$ \textbf{0.32} & \textbf{18.23} $\pm$ \textbf{0.06} & \textbf{34.84} $\pm$ \textbf{0.08} \\
    \bottomrule
    \end{tabular}
\caption{Test errors (\%) on CIFAR-100 and TinyImagenet.
The best results are in bold. The second-best results are underlined. The improvements
are statistically significant
based on the pairwise t-test at 95\% significance level.
}
    \label{tab:cifar100_tiny}
\end{table*}

\subsection{Synthetic Data Classification}
\label{sec:spiral}

Inspired by \cite{ManifoldMIXUP}, this experiment uses the synthetic data set
\textit{Spiral}
to
study the decision boundary learned by the proposed multi-mix.
We generate
1,000 samples.
40\% of them are used for training,
and the rest for testing.
To evaluate the
method's
robustness,
20\% of the training labels
are randomly flipped
(Figure \ref{toy-a}).
We consider manifold mixup with $K=1$
and with
$K=5$
interpolations.
A feed-forward network with eight hidden layers (six units in each layer) and a linear output layer is used.
The interpolation is performed in a shallow layer
(randomly selected from layers 1 and 2).
The network is
$\ell_2$-regularized
and
optimized by the
Adam optimizer \cite{kingma2015adam}, with a learning rate of 0.01.
The
number of epochs is 3,000, and the batch size is 256.

Figure~\ref{fig:spiral} shows the decision boundaries learned without mixup
(Figure~\ref{toy-b}), with manifold mixup
(Figure~\ref{toy-c}),
and with
multi-mix extension of
manifold mixup
(Figure~\ref{toy-d}).
As can be seen,
using multiple interpolations enhances manifold mixup, induces a much smoother decision
boundary, and improves testing accuracy.

\subsection{Classification on CIFAR-100 and TinyImagenet}
\label{sec:cifar}

In this experiment, we
use two standard image classification benchmark datasets: CIFAR-100 \cite{krizhevsky2009learning}
and Tiny-ImageNet \cite{chrabaszcz2017downsampled}.
CIFAR-100 has 100 classes and contains 60,000
32$\times$ 32
color images.
The number of training images and test images are 50,000 and 10,000, respectively.
Tiny-ImageNet contains 200 classes of
$64 \times 64$ images.
Each class has 500 training images and 50 test images.

\subsubsection{Comparison with Various Mixup Variants and Label Smoothing}
\label{sec:part1}

We use three deep networks (PreActResNet18
\cite{he2016identity}, WRN16-8, \cite{zagoruyko2016wide} and ResNeXt29-4-24
\cite{xie2017aggregated})
for CIFAR-100.
Following
\cite{kim2021comixup}, we use the SGD optimizer with an initial learning rate of
0.2, which is decayed by a factor of
0.1 at epochs 100 and 200.
The total number of training epochs is 300.
For
Tiny-ImageNet, we follow the setting in \cite{kim2020puzzle} and use
PreActResNet18 with 1,200 training epochs.
We use the SGD optimizer with an initial learning rate of
0.2, which is decayed by a factor of
0.1 at epochs 600 and 900.
The batch size is 100.
To allow parallelism,
we use PyTorch's multi-process data loading technique \cite{li2020pytorch}
(with 8 loader worker processes).
The experiment is run on a single NVIDIA V6000 GPU.

As discussed in
Section \ref{sec:model},
puzzle-mix can generate more
informative mixup images. Thus, we apply the proposed
multi-mix
(with $K=5$)
on
puzzle-mix.
The use of multi-mix on other mixup variants is shown in Section
\ref{ablation:multiple}.

The proposed method is
compared with the
vanilla classifier (having no mixup)
and various mixup variants
including
(i) input mixup \cite{zhang2018mixup}, (ii) manifold
mixup \cite{ManifoldMIXUP}, (iii) cutout \cite{devries2017CUTOUT}, (iv) cutmix \cite{yun2019cutmix}, (v) FMix \cite{harris2020fmix}, (vi) co-mixup \cite{kim2021comixup}
and
(vii) puzzle-mix \cite{kim2020puzzle}.
Besides, we follow \cite{zhang2018mixup} and include (viii) label smoothing \cite{szegedy2016rethinking}, which
alleviates over-fitting by introducing label noise.

For performance evaluation, as in
\cite{kim2020puzzle,kim2021comixup}, we use the
classification error on
the test set. The experiment is repeated three
times with different random seeds, and then the
average performance is reported.

Table \ref{tab:cifar100_tiny}
shows
the
test errors obtained by the various methods.
As can be seen, multi-mix
outperforms all the other mixup variants.
On CIFAR-100, it achieves improvements of
1.42\%, 0.75\% and 2.29\% over standard puzzle-mix
on networks PreActResNet18, WRN16-8, and
ResNeXt29-4-24,
respectively.
It
also
improves standard puzzle-mix
on TinyImagenet
by 1.68\%.
This demonstrates that leveraging 
multiple interpolations can improve generalization performance.

\subsubsection{Comparison with
Large-Batch Training,
Large-Batch Mixup, and
Batch Augmentation} 
\label{batch_augmentation}

In this experiment,
using the same setup with puzzle-mix as in Section~\ref{sec:part1},
we compare with the baselines discussed in Section~\ref{sec:discuss}
(namely, (i)
batch augmentation
\cite{hoffer2020augment} and (ii) its variant with mixup;
(iii)
large-batch training \cite{keskar2017batch},
and (iv)
large-batch mixup). 

For batch augmentation (and its mixup variant), we keep the original batch size of 100.
As
$K=5$
interpolations are used in multi-mix,
we
run each label-preserving augmentation (random crop, flipping, and rotation)
5 times
for every sample in the batch.
The augmented batch size is thus the same as
that of multi-mix.
Similarly,
the batch size
for large-batch training and large-batch mixup
is
$100 \times 5  = 500$.

\begin{table}[!ht]
\centering 
\renewcommand\arraystretch{1.2}
\begin{tabular}{c|ccc}\toprule
	 & test error (\%) \\\midrule
large-batch training & 33.45 $\pm$ 0.11 	\\
large-batch mixup &  26.54 $\pm$ 0.07 \\
batch augmentation  & 20.91 $\pm$ 0.11 \\
batch augmentation with mixup  & 19.41 $\pm$ 0.08 \\
multi-mix & 	\textbf{18.23 $\pm$ 0.06} \\\bottomrule
\end{tabular}
\caption{Test errors (\%) of various multiple-augmentation strategies 
on CIFAR-100.} 
\label{tab:strategies}
\end{table}

Table \ref{tab:strategies} shows
the
testing errors
on CIFAR-100 using the ResNeXt29-4-24 network.
As can be seen, the performance of
large-batch training
is poor,
as it tends to converge to sharp minima of the training
loss \cite{keskar2017batch}. 
Large-batch mixup alleviates the issue of sharp minima and brings some improvements.
While batch augmentation and its mixup variant lead to improved generalization
performance, multi-mix improves the performance
further
by a larger margin.
This can be explained by the fact that {multi-mix}
can introduce more 
information in the mixup transformation path 
in
one gradient update. Although large-batch mixup and batch augmentation also produce more augmented samples for gradient update, their
generated samples do not include
mixup transformation path information as much
as in {multi-mix}.

\subsection{Large-Scale ImageNet-1K Classification}
\label{sec:imagenet}

In this experiment, we perform
large-scale image classification using the ImageNet data \cite{deng2009imagenet}.
It contains 1.2 million training images and 50,000 validation images from 1,000 classes.
For fair comparison, we follow the training protocol in \cite{kim2021comixup} and report the validation error.

We use the ResNet-50~\cite{he2016deep}, which is trained
for a total of
100 epochs.
As in \cite{kim2021comixup},
these 100 epochs are divided into three stages.
From epochs
0-15,
input images are resized to 160$\times$160.
The initial learning rate is 0.5, and is increased linearly to 1.0 for the first 8
epochs and then decreased linearly to 0.125 until epoch 15.
From epochs
16-40,
images are resized to 352$\times$352. The learning rate is initially set to
0.2, and decreased linearly to 0.02 at
epoch 40.
Finally, for the remaining epochs, the image size is still 352$\times$352, but the
learning rate is
decayed by 0.1 at epochs 65 and 90. SGD is used throughout.
Moreover, only
basic data augmentation strategies, including random crop and random horizon flip,
are used. 
The proposed multi-mix is compared with no mixup,
input mixup, manifold mixup,
cutmix, co-mixup and puzzle-mix.

\begin{table}[!h]
 \centering
\begin{tabular}{c|c}\toprule[1pt]
	 & validation error (\%) \\\midrule
     no mixup & 24.03   \\
     Input mixup & 22.97\\
     Manifold mixup & 23.30\\
     Cutout & 24.10 \\
     Cutmix& 22.92 \\
     FMix & 23.96 \\
     Co-mixup& \underline{22.39}\\
     Puzzle-mix& 22.54  \\
     Label smoothing  & 23.44 \\
     Multi-mix (with
Puzzle-mix) & \textbf{22.29}\\
    \bottomrule
    \end{tabular}
    \caption{Validation error (\%) on ImageNet-1K. The best results are in bold. The second-best results are underlined.
Results of the baselines are from \cite{kim2021comixup}.}
    \label{tab:imagenet}	
\end{table}

Table \ref{tab:imagenet} shows
the
validation
errors.
Co-mixup has the lowest error of 22.39\%
among all baselines,
and outperforms puzzle-mix by 0.15\%.
However, the multi-mix extension of puzzle-mix improves co-mixup by
0.10\%.

\subsection{Weakly-Supervised Object Localization}
\label{sec:wsol}
In this experiment,
we perform weakly-supervised object localization
(WSOL)
\cite{meng2021foreground},
which
uses image-level labels to identify the part containing the target object in an
image from ImageNet
without any pixel-level
supervision.
Following \cite{qin2019rethinking},
we use
the WSOL method
of CAM \cite{zhou2016learning},  which
uses a pre-trained classifier
to obtain the
target object's
bounding box.
Here,
we use the networks trained on ImageNet in Section~\ref{sec:imagenet}
as the pre-trained classifier.

For performance evaluation,
we use the
localization accuracy \cite{kim2021comixup}
(also called
correct localization metric
\cite{cho2015unsupervised}), which is the percentage of images that are correctly localized.
The predicted bounding box
is considered correct if the intersection over union (IoU) \cite{rezatofighi2019generalized}
value between the predicted bounding box and one of the ground-truth bounding boxes is higher than 0.25.
Intuitively, if the network has learned informative representations from the data,
it should perform better on WSOL.

Results are shown in Table \ref{tab:localization}.
Again,
the proposed method outperforms all the baselines. This shows leveraging 
multi-mix helps to learn informative representations in deep networks.

\begin{table}[!h]
 \centering 
\begin{tabular}{c|c}\toprule[1pt]
	 & localization accuracy (\%) \\\midrule
     no mixup &   54.36 \\
     Input mixup &  55.07\\
     Manifold mixup &  54.86\\
     Cutout & 54.24 \\
     Cutmix&  54.91\\
     FMix & 54.57 \\
     Co-mixup&  \underline{55.32}\\
     Puzzle-mix &  55.22 \\
     Label smoothing  & 54.37 \\
     Multi-mix (with
Puzzle-mix) &  \textbf{55.43}\\
    \bottomrule
    \end{tabular}
    \caption{Localization accuracy (\%) in WSOL task. The best results are in bold. The second-best results are underlined. }
    \label{tab:localization}	
\end{table}

\begin{table}[!h]
 \centering
\begin{tabular}{c|cc}\toprule[1pt]
     & random replacement & Gaussian noise \\\midrule
     no mixup &   41.63 & 29.22 \\
     Input mixup &  39.41  & 26.29 \\
     Manifold mixup &  39.72  & 26.79 \\
     Cutout & 41.75 & 29.13 \\
     Cutmix&  46.20 & 27.13\\
     FMix & 41.42 & 28.58 \\
     Co-mixup&  \underline{38.77}  & \underline{25.89} \\
     Puzzle-mix &  39.23 & 26.11 \\
     Label smoothing  & 39.55 & 27.77 \\
     Multi-mix (with
Puzzle-mix) &  \textbf{39.01}  & \textbf{25.53} \\
    \bottomrule
    \end{tabular}
    \caption{Validation errors (\%) on background-corrupted ImageNet datasets. The best results are in bold. The second-best results are underlined. }
    \label{tab:background_corrupted}	
\end{table}

\begin{table*}[!ht] 
 \centering 
    \begin{tabular}{ccccccccccc}
    \toprule[1pt]
& \multicolumn{10}{c}{fine-tuning steps} \\
& 1 & 2 & 3 & 4 & 5& 6& 7& 8 & 9&10 \\\midrule
    no mixup &  28.22 	& 59.67 &	76.59 &	84.48 	& 90.81 &	92.56& 	92.89 	& 93.93 	& 93.85 &	94.43  \\
    Input mixup	&  33.30 	&64.78 	&75.90 	&84.42 	&92.41 	&93.62 	&94.12 	&94.71 &	94.99 &	95.09 \\
    Manifold mixup	& 35.07 	& 67.59 	& 77.96 	& 82.70 	 & 91.71 	& 93.36 	& 94.12 	& 94.62 	& 94.83 	& 95.22  \\
    Cutout 	& 30.70 & 64.34 & 77.99 & 82.91 & 90.95 & 92.59 &  93.45 & 93.97 & 94.12 & 94.49 \\
    Cutmix 	& 39.59 &	\underline{67.99} &	80.92 &	86.47 &	92.93 &	94.04 &	\underline{94.98} &	95.21 &	95.44 &	95.71  \\
    FMix 	& 28.13 & 58.20& 75.34& 82.80&90.13 & 91.96 & 93.10&  93.52 & 93.71 & 94.27\\
    Co-mixup 	& \underline{40.78} & 67.89 & 82.56 & 87.20 & 92.65 & 94.09& 94.90 & 95.19 &  95.43 & 95.81\\
    Puzzle-mix	& 39.37& 70.16 & \underline{82.69} & \underline{87.60} & \underline{93.39} & \underline{94.45} & 94.92 & \underline{95.41} & \underline{95.57} & \underline{95.82}\\
    Label-smoothing & 38.64 &  64.40&  79.31 & 85.48 &  91.86 &  93.37 & 94.17 & 94.68 & 95.01&  95.21\\
    Multi-mix (with Puzzle-mix) & \textbf{45.14} &  \textbf{74.72} & \textbf{84.48} &  \textbf{88.36} & \textbf{94.10}  & \textbf{95.24} &  \textbf{95.94} & \textbf{96.08} & \textbf{96.23} & \textbf{96.49}\\
    \bottomrule
    \end{tabular}
    \caption{Localization accuracy (in \%)
    with different numbers of fine-tuning steps
	 on \emph{CUB 200-2011}. The best results are in bold. The second-best results are underlined. }
    \label{tab:transfer}
\end{table*}

\subsection{Robustness to Corruption}
\label{sec:corrupt}

In this section, we evaluate the robustness
to
images with
corrupted background
of
the various trained ResNet-50 models in Section~\ref{sec:imagenet}.
As in
\cite{kim2021comixup},
the ImageNet validation set images are used. Their
backgrounds
are corrupted
by either
(i) random replacement with
another image from the validation set,
or (ii) addition of Gaussian noise from $\mathcal{N}(0, 0.1^2)$.

Table \ref{tab:background_corrupted}
shows the
	 validation errors.
The proposed puzzle-mix
with multiple interpolations
shows good robustness against both kinds of background
corruption.
For random replacement, the use of multiple interpolations improves puzzle-mix,
and is competitive with co-mixup.
For Gaussian-noise corruption, the proposed method outperforms all mixup baselines.
This supports the intuition that multiple interpolations help to induce more
robust representations.

\subsection{Transfer Learning to Fine-Grained CUB 200-2011}
\label{sec:xfer}
In this experiment, we study the transferability of the various ResNet-50 models trained in Section~\ref{sec:imagenet}.
Each model
is transferred
to the task of object localization
on the fine-grained CUB 200-2011 dataset
\cite{wah2011caltech}, by
fine-tuning
the last ResNet block and fully-connected
layer
for a maximum of
10 epochs
(using
the Adam optimizer with a learning rate of 0.0001).
We use the
localization accuracy
for performance evaluation.
The binarization threshold is set to 0.75 as in \cite{Wu_2020_CVPR}.

Table \ref{tab:transfer} shows
the localization accuracies when the number of fine-tuning steps is varied from 1 to 10. Note that
using more fine-tuning steps improves performance.
As can be seen, the network pretrained using multiple interpolations takes fewer
fine-tuning steps and has better localization performance
than networks pretrained by the other mixup methods.

\begin{figure*}[!t]
\centering
\subcaptionbox{\label{0}no mixup.}
{\includegraphics[width=0.18\textwidth]
{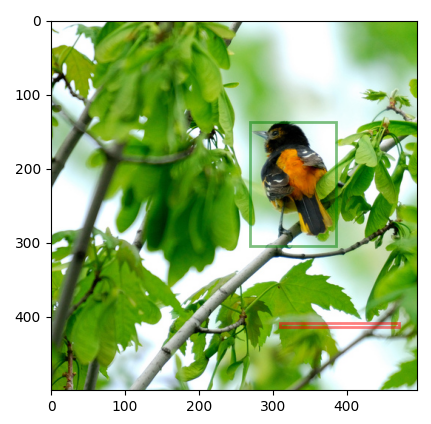}}
\subcaptionbox{\label{1}Input mixup.}
{\includegraphics[width=0.18\textwidth]{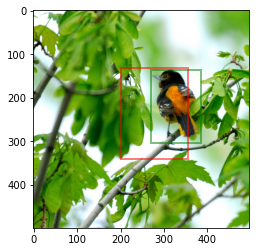}}
\subcaptionbox{\label{2}Manifold mixup.}
{\includegraphics[width=0.18\textwidth]{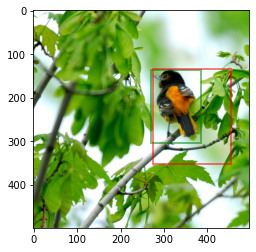}}
\subcaptionbox{\label{3}Cutmix.}
{\includegraphics[width=0.18\textwidth]{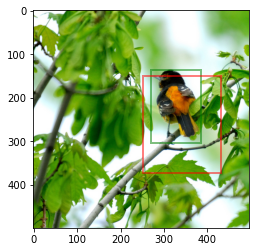}}
\subcaptionbox{\label{4}Cutout.}
{\includegraphics[width=0.18\textwidth]{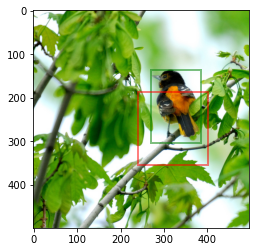}}
\subcaptionbox{\label{5}FMix.}
{\includegraphics[width=0.18\textwidth]{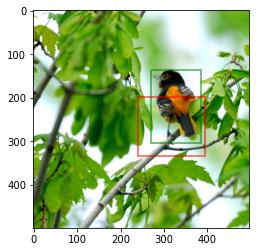}}
\subcaptionbox{\label{6}Co-mixup.}
{\includegraphics[width=0.18\textwidth]{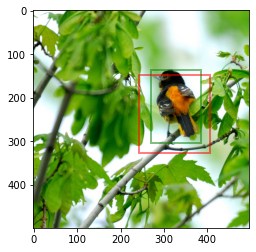}}
\subcaptionbox{\label{7}Puzzle-mix.}
{\includegraphics[width=0.18\textwidth]{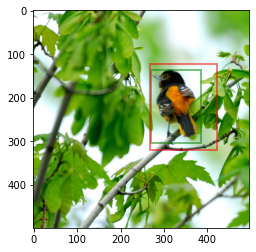}}
\subcaptionbox{\label{8}Label-smoothing.}
{\includegraphics[width=0.18\textwidth]{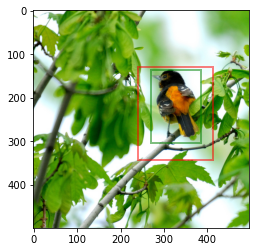}}
\subcaptionbox{\label{9}\makecell[c]{Multi-mix\\(with Puzzle-mix)}.}
{\includegraphics[width=0.18\textwidth]{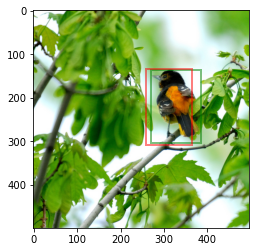}}
\caption{Example localization results
	 on \emph{CUB 200-2011}
from networks pretrained with different
mixup methods. Red: Predicted bounding box; Green: Ground-truth bounding box.}
\label{fig:localization}
\end{figure*}

Figure \ref{fig:localization} shows
examples produced by the networks pretrained with different mixup methods with only
one fine-tuning step.
As can be seen, the predicted bounding box by the proposed method localizes the
target object with higher accuracy.
This supports that networks augmented by the proposed method are more informative.

\begin{table}[!t]
 \centering
\begin{tabular}{c|c}\toprule[1pt]
	 & test error (\%) \\\midrule
     no mixup &   4.84 \\
     Input mixup &  3.91 \\
     Manifold mixup &  3.67 \\
     Cutout & 3.76 \\
     Cutmix&  4.36 \\
     FMix & 4.10 \\
     Co-mixup&  \underline{3.54}\\
     Puzzle-mix &  3.70 \\
     Label smoothing  & 4.72 \\
     Multi-mix (with
Puzzle-mix) &  \textbf{3.41}\\
    \bottomrule
    \end{tabular}
    \caption{Test error (\%) on the Google commands dataset. The best results are in bold. The second-best results are underlined. }
    \label{tab:speech}
\end{table}

\subsection{Classification on Speech Data}
\label{sec:speech}
In this experiment, we perform speech classification on the Google commands
dataset\footnote{{https://ai.googleblog.com/2017/08/launching-speech-commands-dataset.html}}
\cite{warden2018speech}.
This includes 65,000 utterances from 30 categories (e.g., ``On", ``Off", ``Stop", and
``Go").
Following \cite{zhang2018mixup},
the raw waveform utterances
are preprocessed
by short-time Fourier transform \cite{durak2003short}
to extract
normalized spectrograms.
We then
use the VGG-11 network \cite{simonyan2014very}
and
perform various mixup methods at the spectrogram level.
The network is trained by
SGD
for 30 epochs.
The initial learning rate is $0.003$ and is decayed by a factor of 10 every
10 epochs. This dataset has a standard training/validation/test split.
Following \cite{kim2021comixup}, we select the best
model based on the validation error and report the corresponding test error.

Table \ref{tab:speech} shows the test errors.
As can be seen, the proposed method again outperforms all the baselines.

\begin{figure*}[!t]
 \centering
  \includegraphics[width=0.85\linewidth]{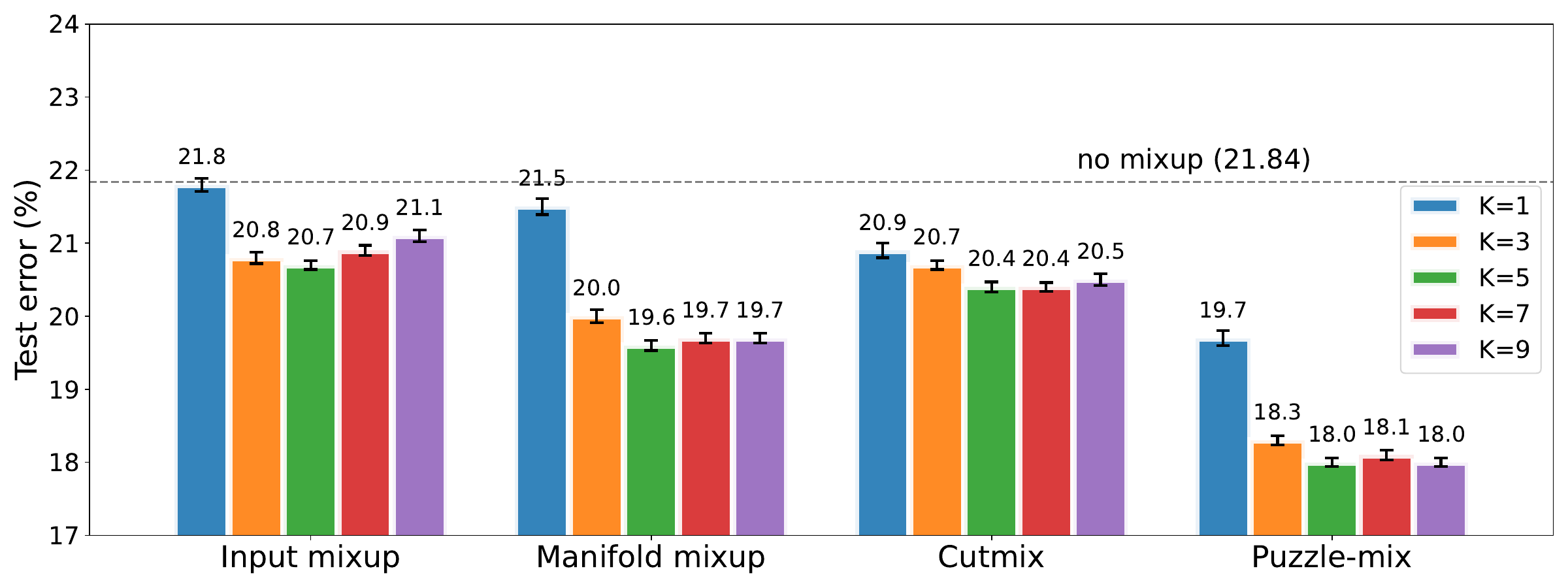}
  \captionof{figure}{Test errors of
 the proposed \textit{multi-mix} extensions on
  various mixup strategies with different $K$'s. For comparison, the performance with no mixup is shown by the dotted line.
  }
  \label{fig:acc_varyingK}
 \end{figure*}

\subsection{Calibration Analysis}
\label{sec:calib}
Deep networks trained with mixup are often better calibrated than models
trained without mixup
\cite{thulasidasan2019mixup}.
A well-calibrated model is not over-confident and produces softmax scores that are close to
the actual likelihoods of correctness.
In this experiment, we study the ResNet-50 models trained on the ImageNet
in Section~\ref{sec:imagenet}.

Performance evaluation is  based on
the  popularly-used
expected calibration error (ECE)~\cite{guo2017calibration,kim2021comixup}.
Specifically, predictions
from all test samples
are grouped into $M$ bins (each of size
$1/M$).
Following \cite{kim2021comixup}, we used $M=10$.
Let
$B_{m}$ be the set of samples whose largest softmax scores fall in the interval $I_{m}=\left(\frac{m-1}{M}, \frac{m}{M}\right]$.
The average accuracy of bin $B_{m}$ is
$$\operatorname{acc}(B_{m})=\frac{1}{|B_{m}|} \sum_{i \in B_{m}}
\mathbf{1}(\hat{y}_{i}=y_{i}),$$
where $\hat{y}_{i}$ and $y_{i}$ are the predicted and true class labels for sample $i$, respectively.
Similarly,
the average confidence of bin $B_{m}$ is defined as:
$$\operatorname{conf}(B_{m})=\frac{1}{|B_{m}|} \sum_{i \in B_{m}} \hat{p}_{i},$$
where $\hat{p}_{i}$ is the
logistic output
of sample $i$.
ECE is then defined as:
$$\mathrm{ECE}=\sum_{m=1}^{M}
\frac{|B_{m}|}{n}|\operatorname{acc}(B_{m})-\operatorname{conf}(B_{m})|.$$
A well-calibrated model should have a small ECE.

\begin{table}[!h]
 \centering
\begin{tabular}{c|c}\toprule[1pt]
	 & ECE (\%) \\\midrule
     no mixup &   5.93 \\
     Input mixup &  \textbf{1.18} \\
     Manifold mixup &  1.70 \\
     Cutout & 5.72 \\
     Cutmix&  4.31 \\
     FMix & 5.89 \\
     Co-mixup&  {2.17}\\
     Puzzle-mix &  2.06 \\
     Label smoothing  & 5.39 \\
     Multi-mix (with
Puzzle-mix) &  \underline{1.46}\\
    \bottomrule
    \end{tabular}
    \caption{ECE (\%) of the ResNet-50 models trained on ImageNet. The best results are in bold. The second-best results are underlined. }
    \label{tab:ECE}
\end{table}

Table \ref{tab:ECE} shows the ECE results.
As can be seen,
the network trained with
the proposed method is better calibrated than most baselines (except input
mixup).
While input mixup leads to the lowest ECE, its classification error on the same
task is high (see
Table~\ref{tab:imagenet}).
Thus, the proposed method achieves very competitive ECE performance while having the lowest classification error.

\begin{figure}[!t]
  \centering
  \includegraphics[width=0.85\linewidth]{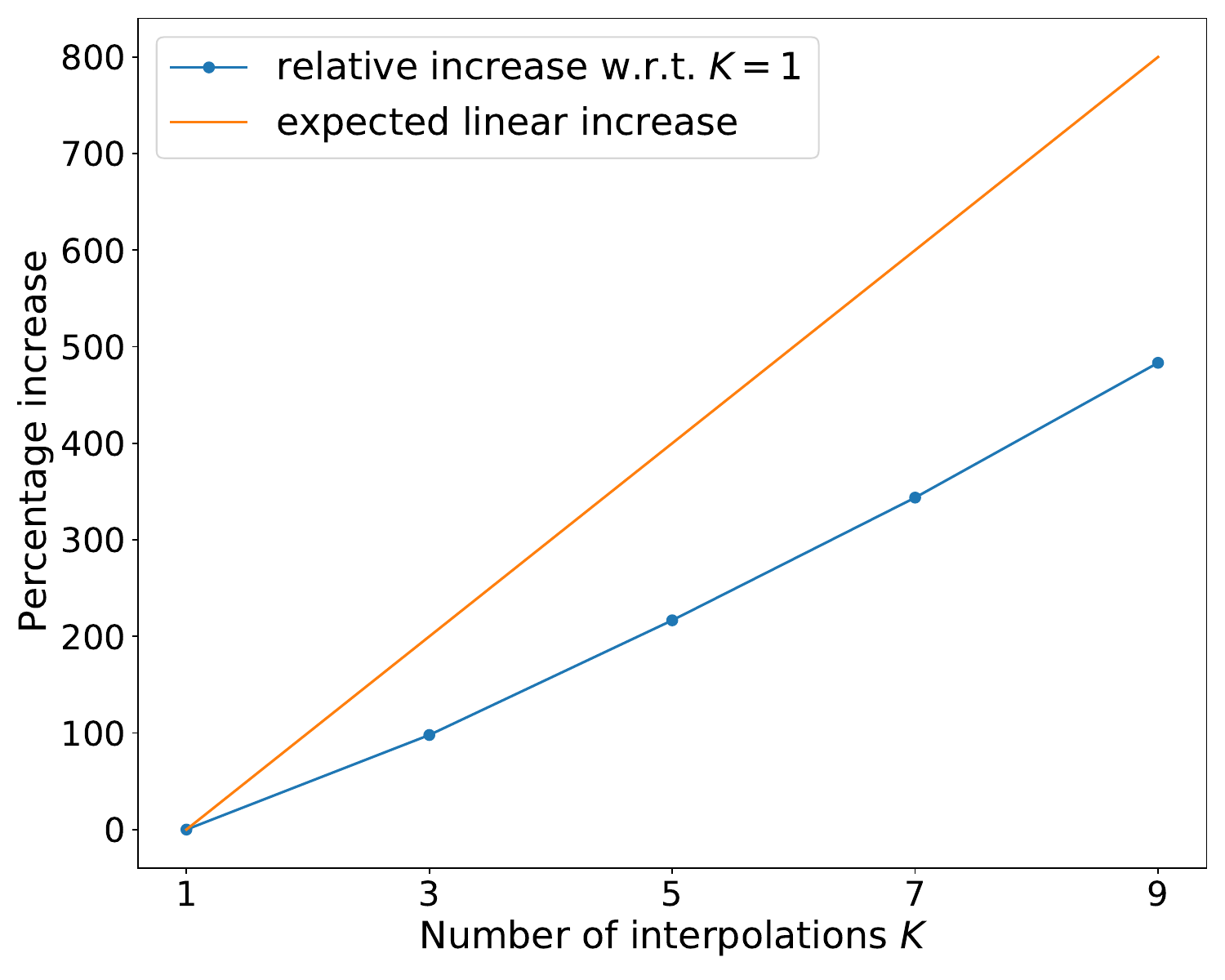}
  \caption{Percentage increase in per-epoch training time with $K$ (linear
  scaling is shown in orange).}
  \label{fig:acceleration}
\end{figure}

\subsection{Ablation Study:
Effect of $K$}
\label{ablation:multiple}

In this experiment, we study the effect of $K$, the number of interpolations. We compare the
training
time of puzzle-mix
(with $K=1$ interpolation)
and its multi-mix version
(with
$K=3,5,7,9$
interpolations)
on \emph{CIFAR100}.
We use the setup in Section \ref{sec:cifar}.
Besides puzzle-mix, we also study the multi-mix extensions of input mixup, manifold mixup and cutmix. 
The network used is ResNeXt29-4-24, which is trained in the same manner
as in section~\ref{sec:cifar}. 
The experiment is repeated three times, and both the mean test error and its
standard deviation are reported.

Figure \ref{fig:acc_varyingK}
shows the test error
with different
$K$'s.
As can be seen, the use of multiple interpolations benefits all mixup variants to
different extents. Improvements in input mixup and cutmix are
relatively small.
This may be due to that their
mixing results
may lose
some object information in the source images,
and cutmix causes sharp mixing-box boundaries
(see Fig \ref{fig:mixup_example}).
For puzzle-mix,
the performance improves with
larger $K$.
However, a large $K$ is not necessary. In practice, we set $K$ to
5.
 
Figure~\ref{fig:acceleration}
shows the
relative increase in
per-epoch
training
time
of multi-mix (with puzzle-mix)
over standard puzzle-mix
with different values of $K$.
As can be seen, the
training time increases
with $K$, though
sublinearly because of the use of parallelism. 

Finally, we
provide empirical evidence for the reduction in gradient variance as discussed in
Section~\ref{sec:theory}.
Note that directly computing the gradient variance
using (\ref{eq:gvar})
can be difficult, as it requires computing both $\E[\| \tilde{\vg} \|^2]$ and
$\E[\tilde{\vg}]$.
However, for any $K$, we have $\E[\tilde{\vg}] = \nabla \L_{{mixup}}$
(section \ref{sec:theory})
which does not
depend on $K$. Hence,
a larger
$\E[\| \tilde{\vg} \|^2]$
indicates a larger gradient variance $\Var[\tilde{\vg}]$.
As such, following~\cite{hoffer2020augment},
we use the expectation of the
squared norm
$\| \tilde{\vg} \|^2$
as a surrogate to compare $\Var[\tilde{\vg}]$.
We estimate the gradient norm in each batch and average over the whole epoch. 

\begin{figure}[!t]
 \centering
  \includegraphics[width=0.99\linewidth]{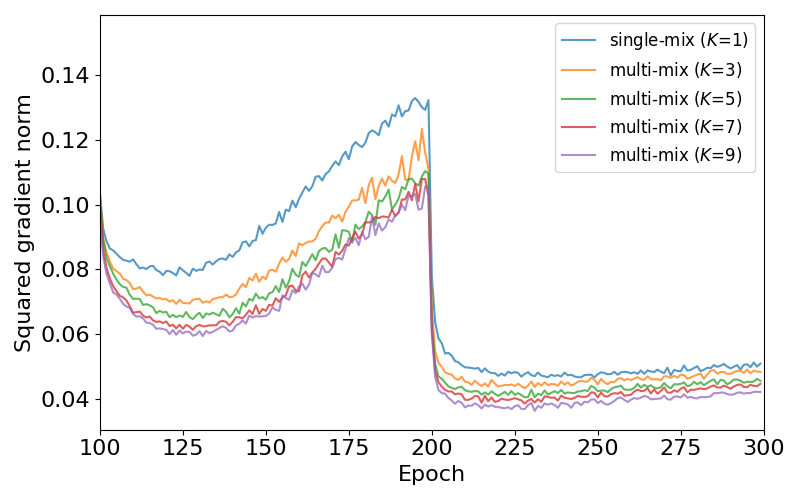}
\captionof{figure}{$\ell_2$-norm of gradients versus number of training epochs
on CIFAR-100.
Note that the drop at epoch 200 in Figure \ref{fig:gradnorm} is a result of the
learning rate decay by a factor of 0.1.}
  \label{fig:gradnorm}
 \end{figure}

Figure \ref{fig:gradnorm} shows
the squared $\ell_2$-norm
of the stochastic
gradient with the number of training epochs.
As can be seen,
the use of a large $K$ reduces the gradient norm,
thus
supporting the theoretical analysis
in Proposition~\ref{prop:var1}.

\section{Conclusion}
In this work, we propose a simple yet powerful extension of mixup and its variants
with the use of multiple interpolations.
Theoretical analysis shows that it can reduce the variance of stochastic gradients,
leading to better optimization.
Extensive experiments demonstrate that this
can
improve generalization,
induce smoother decision boundaries,
and learn more informative representations. Moreover,
the obtained network is more robust against data corruption, more transferable, and better calibrated.
Hence, this leveraging of mixup order relationships offers an easy and effective
tool to
improve the mixup family.

\begin{appendix}
\section{Proofs}

\subsection{Proof of Proposition~\ref{prop:var1}} \label{sec:proof}

Let the variance of $\vg((\x,\y), (\x', \y'), \lambda)$ be $$\Var[\vg((\x,\y), (\x', \y'), \lambda)]
= \E[\| \vg((\x,\y), (\x', \y'), \lambda)- \vg \|^2],$$ where
\[ \vg
= \E_{(\x,\y), (\x', \y')} \E_{\lambda_k}[\vg((\x,\y), (\x', \y'), \lambda_k)]. \]
For simplicity of notations,
$\lambda_k$ is always sampled from
$Beta(\alpha,\alpha)$
in the sequel.
Since we have two sources of stochasticity (sample pairs and $\lambda$),
and they are independent of each other, we can decouple the variance
as:
\begin{eqnarray*}
\lefteqn{\Var[\vg((\x,\y), (\x', \y'), \lambda_k)]} \\
& =  & \E_{(\x,\y), (\x', \y')} \E_{\lambda_k}[\| \vg((\x,\y), (\x', \y'), \lambda_k) - \vg \|^2] \notag\\
&= & \E_{(\x,\y), (\x', \y')} \Var_{\lambda_k}[\vg((\x,\y), (\x', \y'), \lambda_k)] \\
& & + \Var_{(\x,\y), (\x', \y')}[\bar{\vg}_1((\x,\y), (\x', \y'))],
\end{eqnarray*}
where
\begin{eqnarray*}
& & \Var_{\lambda_k}[\vg((\x,\y), (\x', \y'), \lambda_k)] \\
& = & \E_{\lambda_k}[\| \vg((\x,\y), (\x', \y'),
\lambda_k) - \bar{\vg}_1((\x,\y), (\x', \y')) \|^2], \\
& & \Var_{(\x,\y), (\x', \y')}[\bar{\vg}_1((\x,\y), (\x', \y'))] \\
& =  & \E_{(\x,\y), (\x', \y')}[\| \bar{\vg}_1((\x,\y), (\x', \y')) - \vg \|^2],
\end{eqnarray*}
and
$\bar{\vg}_1((\x,\y), (\x', \y')) =  \mathop{\mathbb{E}}_{\lambda_k }[\vg((\x,\y), (\x', \y'), \lambda_k)]$.
In other words,
$\Var_{\lambda_k }$
is the variance due to $\lambda_k$,
and $\Var_{(\x,\y), (\x', \y')}$ is the variance due to sample pair $\{(\x,\y), (\x', \y')\}$.

Similarly, we can also obtain that:
\begin{align*}
& \Var[\vg((\x,\y), (\x', \y'), \lambda_k)] \\
= & \E_{\lambda_k} \Var_{(\x,\y), (\x', \y')} [\vg((\x,\y), (\x', \y'), \lambda_k)]  + \Var_{\lambda_k}[\bar{\vg}_2(\lambda_k)],
\end{align*}
where
\begin{eqnarray*}
& & \Var_{(\x,\y), (\x', \y')}[\vg((\x,\y), (\x', \y'), \lambda_k)] \\
& = &  \E_{(\x,\y), (\x', \y')}[\| \vg((\x,\y), (\x', \y'), \lambda_k) - \bar{\vg}_2(\lambda_k) \|^2,\\
& & \Var_{\lambda_k }[\bar{\vg}_2(\lambda_k)] \\
& = & \E_{\lambda_k}[\| \bar{\vg}_2(\lambda_k) - \vg \|^2],
\end{eqnarray*}
and $\bar{\vg}_2(\lambda_k) = \E_{\x, \x'}[\vg((\x,\y), (\x', \y'), \lambda_k)]$.

As
$\Var_{\lambda_k }[\vg((\x,\y), (\x', \y'), \lambda_k)] $
and
$\Var_{(\x,\y), (\x', \y')}[\vg((\x,\y), (\x', \y'), \lambda_k)]$ are both non-negative,
we have
\begin{align}
\Var[\vg((\x,\y), (\x', \y'), \lambda_k)] &\ge \Var_{(\x,\y), (\x', \y')}[\bar{\vg}_1((\x,\y), (\x', \y'))] \ge 0,\notag\\
\Var[\vg((\x,\y), (\x', \y'), \lambda_k)] &\ge
\Var_{\lambda_k}[\bar{\vg}_2(\lambda_k)] \ge 0.
\label{eq:app1}
\end{align}
Now we are ready to study the variance of $\tilde{\vg}$ in
(\ref{eq:grad_multimix}).
\begin{eqnarray*}
\lefteqn{\Var[\tilde{\vg}]}\\
& = & \E\left[\left\| \frac{1}{B} \sum_{\substack{(\x,\y) \in \gB,\\ (\x', \y') \in \gB}}
\frac{1}{K} \sum_{k=1}^K \vg((\x,\y), (\x', \y'), \lambda_k) - \vg \right\|^2\right] \\
& = & \frac{1}{B^2} \E\left[\left\| \sum_{\substack{(\x,\y) \in \gB,\\ (\x', \y') \in \gB}} \left(\frac{1}{K}
\sum_{k=1}^K \vg((\x,\y), (\x', \y'), \lambda_k) - \vg \right) \right\|^2\right] \\
& = & \frac{1}{B^2} \left( \sum_{\substack{(\x,\y) \in \gB,\\ (\x', \y') \in \gB}}  \E\left[\left\| \frac{1}{K}
\sum_{k=1}^K \vg((\x,\y), (\x', \y'), \lambda_k) - \vg \right\|^2 \right] \right. \\
&& \left. \!+\! \sum_{\substack{\substack{(\x,\y) \in \gB,\\ (\x', \y') \in \gB},\\ \substack{(\u, \v) \in \gB,\\ (\u', \v') \in \gB},\\ \x \ne \u, \x' \ne \u'}} \E[\mathbf{p}((\x,\y),(\x',\y'))^{\top} \mathbf{p}((\u,\v),(\u',\v'))] \right),
\end{eqnarray*}
where $\mathbf{p}((\x,\y), (\x', \y'))= \frac{1}{K} \sum\nolimits_{k=1}^K \vg((\x,\y), (\x', \y'),
\lambda_k) - \vg$.
For the first term,
we have:
\begin{eqnarray*}
\lefteqn{\E\left[\left\| \frac{1}{K} \sum\nolimits_{k=1}^K \vg((\x,\y), (\x', \y'), \lambda_k) - \vg
\right\|^2 \right]} \\
& = & \frac{1}{K^2} \left( \sum_{k=1}^K \E[\| \vg((\x,\y), (\x', \y'), \lambda_k) - \vg \|^2]
\right. \\
& & \left. + \sum_{k, k'=1, k \ne k'}^K \E[(\vg((\x,\y), (\x', \y'), \lambda_k) -
\vg)^{\top} \right. \\
&& \left. (\vg((\x,\y), (\x', \y'), \lambda_{k'}) - \vg)] \right) \\
& = & \frac{1}{K} ( \Var[\vg((\x,\y), (\x', \y'), \lambda_k)] \\
& & + (K-1) \Var_{(\x,\y), (\x', \y')}[\bar{\vg}_1((\x,\y), (\x', \y'))]),
\end{eqnarray*}
where the division by $K$ comes from using $K$ interpolations.
For the second term,
since we
have different sample pairs, we have:
\begin{eqnarray*}
\lefteqn{\E[\mathbf{p}((\x,\y), (\x', \y'))^{\top} \mathbf{p}((\u, \v), (\u', \v'))]}\\
& = &
\E_{\lambda_k } \left(\frac{1}{K} \sum\nolimits_{k\!=\!1}^K \bar{\vg}_2(\lambda_k) \!-\! \vg \right)^{\top} \left(\frac{1}{K} \sum\nolimits_{k\!=\!1}^K \bar{\vg}_2(\lambda_k) \!-\! \vg \right) \\
& = & \frac{1}{K^2} \left(\sum_{k=1}^K \E_{\lambda_k }[\| \bar{\vg}_2(\lambda_k) - \vg \|^2] \right. \\
&& + \left. \sum_{k, k'=1, k\ne k'}^K \E_{\lambda_k, \lambda_{k'} }[(\bar{\vg}_2(\lambda_k) - \vg)^{\top} (\bar{\vg}_2(\lambda_{k'}) - \vg) ] \right) \\
&   = & { \frac{1}{K} \Var_{\lambda_k }[\bar{\vg}_2(\lambda_k)]}.
\end{eqnarray*}
Adding these two terms together, we have
\begin{eqnarray}
\lefteqn{\Var[\tilde{\vg}]} \nonumber \\
& =  & \frac{1}{K} \left( \frac{1}{B} \Var[\vg((\x,\y), (\x', \y'), \lambda_k)]
 +  \frac{B\!-\!1}{B} \Var_{\lambda_k }[\bar{\vg}_2(\lambda_k)] \right) \nonumber \\
 && +  \left(1\!-\!\frac{1}{K} \right) \cdot \frac{1}{B} \Var_{(\x,\y), (\x', \y')}[\bar{\vg}_1((\x,\y), (\x', \y'))]
\label{eq:appaa}
\\
& =& \frac{1}{B} \Var_{(\x,\y), (\x', \y')}[\bar{\vg}_1((\x,\y), (\x', \y'))] \\
& & +
\frac{1}{K} \left( \frac{1}{B} \Var[\vg((\x,\y), (\x', \y'), \lambda_k)] \right. \notag \\
&& \left. + \frac{B\!-\!1}{B} \Var_{\lambda_k }[\bar{\vg}_2(\lambda_k)] \right. \notag \\
& & \left. - \frac{1}{B}
\Var_{(\x,\y), (\x', \y')}[\bar{\vg}_1((\x,\y), (\x', \y'))] \right). \nonumber
\end{eqnarray}
Consider
$\frac{1}{B} \Var[\vg((\x,\y), (\x', \y'), \lambda_k)]  +  \frac{B-1}{B} \Var_{\lambda_k
}[\bar{\vg}_2(\lambda_k)]$ in
the second term on the RHS.
Using (\ref{eq:app1}), we have:
\begin{eqnarray*}
\lefteqn{\frac{1}{B} \Var[\vg((\x,\y), (\x', \y'), \lambda_k)]  +  \frac{B-1}{B} \Var_{\lambda_k }[\bar{\vg}_2(\lambda_k)]} \nonumber \\
& \ge &  \frac{1}{B} \Var[\vg((\x,\y), (\x', \y'), \lambda_k)] \\
& \ge & \frac{1}{B} \Var_{(\x,\y), (\x', \y')}[\bar{\vg}_1((\x,\y), (\x', \y'))] \geq 0.
\end{eqnarray*}
Hence,
when $K$ increases,
the
second term decreases,
which finishes the
proof.

\subsection{Proof of Proposition~\ref{prop:var}} \label{sec:proof_2}
Based on Proposition~\ref{prop:var1}, we can similarly compute the variance of $\tilde{\vg}'$ from (\ref{eq:appaa})
by setting $K'=1$ and $B'=KB$ for $\tilde{\vg}'$, as follows:
\begin{eqnarray*}
\lefteqn{\Var[\tilde{\vg}']}\\
&= &\frac{1}{K'B'} \big(  \Var[\vg((\x,\y), (\x', \y'), \lambda_k)] \\
& & + (K'\!-\!1) \Var_{(\x,\y), (\x', \y')}[\bar{\vg}_1((\x,\y), (\x', \y'))] \\
&&+  (B'-1)\Var_{\lambda_k }[\bar{\vg}_2(\lambda_k)] \big) \\
&= & \frac{1}{KB} \left( \Var[\vg((\x,\y), (\x', \y'), \lambda_k)] \right. \\
& & \left. + (KB-1)\Var_{\lambda_k}[\bar{\vg}_2(\lambda_k)] \right).
\end{eqnarray*}
to compare the variances of $\tilde{\vg}$ and $\tilde{\vg}'$, obviously we have
\begin{eqnarray*}
\lefteqn{\Var[\tilde{\vg}] \le \Var[\tilde{\vg}']} \\
& \Leftrightarrow & \Var_{\x,
\x'}[\bar{\vg}_1((\x,\y), (\x', \y'))] \le B \cdot \Var_{\lambda_k}[\bar{\vg}_2(\lambda_k)].
\end{eqnarray*}
Note that this does not depend on the choice of $K$ as long as $K \ge 1$ (which is trivial).
In other words, we can let $B_0$ such that
\begin{eqnarray*}
B & \ge & \frac{\Var_{(\x,\y), (\x', \y')}[\bar{\vg}_1((\x,\y), (\x', \y'))]}{\Var_{\lambda_k}[\bar{\vg}_2(\lambda_k)]} \\
& =& \frac{\Var_{(\x,\y), (\x', \y')} \mathbb{E}_{\lambda_k }[\vg((\x,\y), (\x', \y'), \lambda_k)]}{\Var_{\lambda_k}\mathbb{E}_{(\x,\y), (\x', \y')}[\vg((\x,\y), (\x', \y'), \lambda_k)]} = B_0,
\end{eqnarray*}
which finishes the proof.

\end{appendix}

\bibliography{myref}
\bibliographystyle{ieeetr}

\newpage

\appendices

\onecolumn

\end{document}